\documentclass[letterpaper, 10 pt, conference]{ieee/ieeeconf}  

\IEEEoverridecommandlockouts                              
\usepackage{graphics} 
\usepackage{epsfig} 
\usepackage{mathptmx} 
\usepackage{times} 
\usepackage{amsmath} 
\usepackage{amssymb}  
\usepackage{xcolor}
\usepackage{color}
\usepackage[linesnumbered,ruled,vlined]{algorithm2e}
\usepackage{algorithmic}
\usepackage{subfigure}
\usepackage{multirow}
\usepackage{float}
\usepackage{algorithm2e}
\usepackage{booktabs}
\usepackage[utf8x]{inputenc}
\usepackage{wrapfig}

\renewcommand{\baselinestretch}{1}

\SetKwComment{Comment}{$\triangleright$\ }{}

\title{\LARGE \bf
Harmonious Sampling for Mobile Manipulation Planning
}

\author{Mincheul Kang$^{1}$, Donghyuk Kim$^{1}$ and Sung-Eui Yoon$^{2}$
\thanks{{$^1$}Mincheul Kang ({\tt\small mincheul.kang@kaist.ac.kr}) and
{$^1$}Donghyuk Kim ({\tt\small donghyuk.kim@kaist.ac.kr}) are with the School of Computing, and
{$^2$}Sung-Eui Yoon (Corresponding author, {\tt\small sungeui@kaist.edu}) is with the Faculty of School of Computing, KAIST at Daejeon, Korea 34141}
}

\newcommand{\Skip}[1]{}
\renewcommand{\paragraph}[1]{{\bf {#1}}}  

\def\HiLi{\leavevmode\rlap{\hbox to 
\hsize{\color{yellow!50}\leaders\hrule height .8\baselineskip depth .5ex\hfill}}}

\begin{document}

\maketitle
\thispagestyle{empty}
\pagestyle{empty}

\newtheorem{thm}{Theorem}
\newtheorem{lem}[thm]{Lemma}
\newtheorem{col}[thm]{Corollary}

\begin{abstract}
Mobile manipulation planning commonly adopts a decoupled approach that performs
	planning separately on the base and the manipulator.
	While this approach is fast, it can generate sub-optimal
	paths.
	Another direction is a coupled
	approach jointly adjusting the base and manipulator in a
	high-dimensional configuration space.
This coupled approach addresses sub-optimality and incompleteness of the
decoupled approach, but has not been widely used due to its excessive
computational overhead.
Given this trade-off space, we present a simple, yet effective mobile manipulation
	sampling method, harmonious sampling, to perform the coupled approach
	mainly
	in difficult regions,
	where we need to simultaneously maneuver the base and
	the manipulator.
Our method identifies such difficult regions through a low-dimensional base
space by utilizing
	a reachability map given the target
	end-effector pose and narrow passage detected by generalized
	Voronoi diagram. 
	For the rest of simple regions, 
	we perform sampling mainly on the base configurations with a predefined
	joint configuration, accelerating the planning process.   We compare
	our method with the decoupled and coupled
	approaches in six different problems with varying difficulty.  Our
	method shows meaningful improvements experimentally in terms of time to
	find an initial solution (up to 5.6 times faster) and final solution cost
	(up to 17$\%$ lower) over the decoupled approach, especially in
	difficult scenes with narrow space.  We also demonstrate these benefits
with a real, mobile Hubo robot.  \end{abstract}

\section{INTRODUCTION}
\label{sec:1}

Mobile manipulation planning has been attracted much attention in recent years
since the rise of demands for autonomous indoor services, e.g., households
tasks or delivery services.  Mobile manipulators usually equipped with wheels
on a base for the mobility perform various manipulation tasks, e.g., a pick-and-place 
problem, with robotic arms.

For the mobile manipulation planning problem, a target end-effector pose is
given.  Given the pose, we first compute its base configuration by utilizing a
reachability inversion~\cite{vahrenkamp2013robot} and then set joints of the
manipulator based on inverse kinematics for computing a goal
configuration in the joint configuration space (C-space) consisting of the base
and manipulator.
Unfortunately, this joint C-space can be high-dimensional, and thus tends to
lead a slow planning process.

Therefore, a decoupled approach that plans separately for the base and the
manipulator has been commonly used in practice, since this decoupled approach
works relatively fast thanks to planning in separated and thus
lower-dimensional spaces~\cite{chitta2012perception}.
Nonetheless, this decoupled approach may find sub-optimal results or let the
manipulator to place in a narrow space due to the fixed base used for
manipulation planning. 
This is mainly caused by its decoupled planning and sampling that does not
consider the joint planning between the base and manipulator.

\begin{figure}[t]
	\vspace{0.25cm}
	\centering
	\includegraphics[width=\columnwidth]{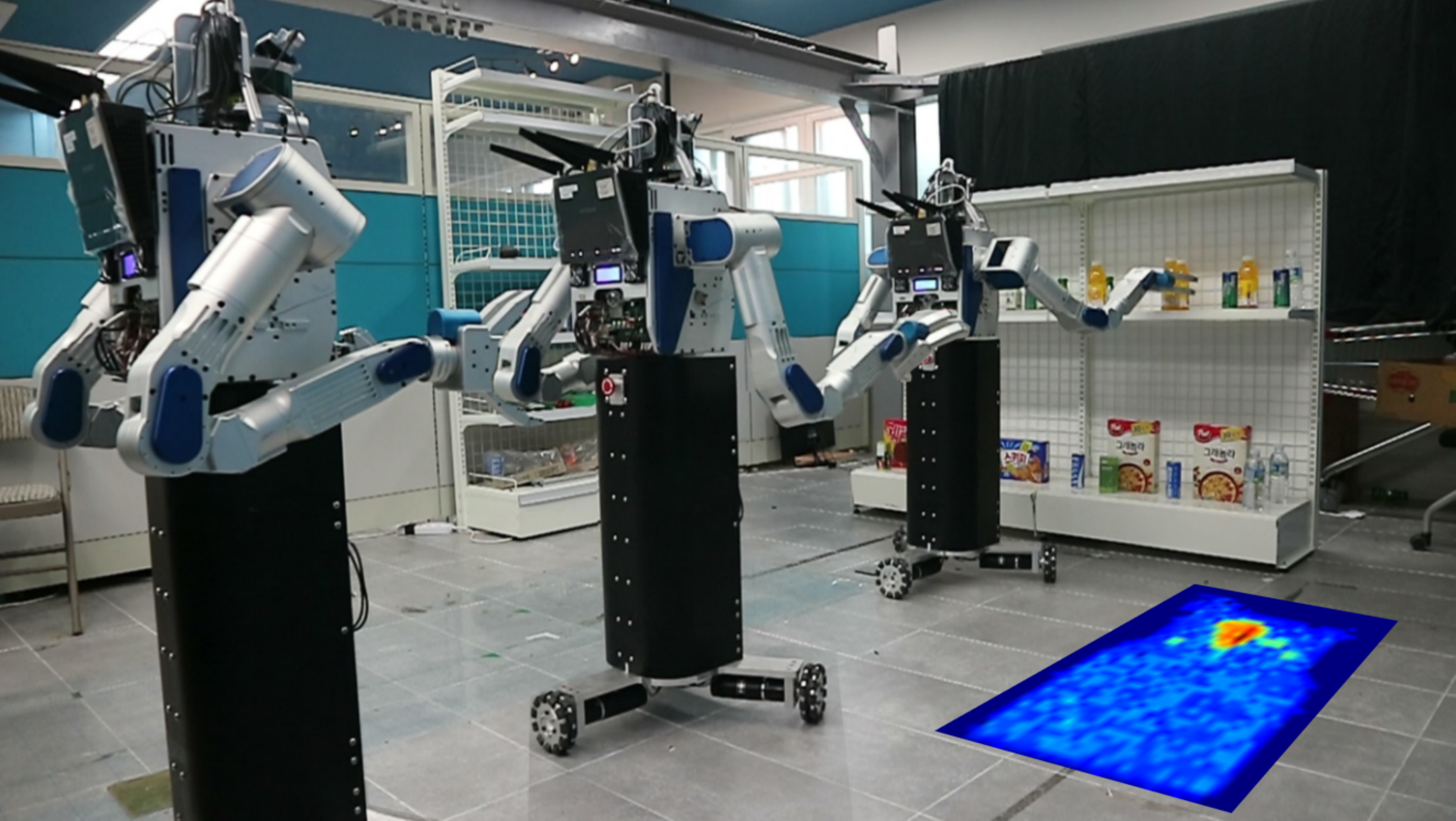}
	\caption{
		This figure shows a sequence of the mobile Hubo robot
		grasping the yellow beverage.  
		The heat map on the bottom right represents a sample
		density in terms of the 2D floor projected from samples
		generated from our harmonious sampling in the joint C-space
		consisting of the base and the manipulator.
		Samples are distributed intensively near the goal configuration, resulting in efficient and effective exploration of solution paths.
	}
	\label{fig:main}
	\vspace{-0.6cm}
\end{figure}

\noindent
\textbf {Main contributions.} 
In this work, we present a simple, yet effective mobile manipulation sampling
method, harmonious sampling, for optimal mobile manipulation planning, to
efficiently explore a high-dimensional C-space consisting of both base
and manipulator (Fig.~\ref{fig:main}).
Our harmonious sampling adjusts the sampling space for the base and the
manipulator, and guides to sample more on difficult regions where we need to
simultaneously change the configurations of both base and manipulator
(Sec.~\ref{sec:4_harmonious_sampling}).
We identify such regions based on a low-dimensional base space, especially,
manipulation regions that are computed by considering the target end-effector
pose and obstacles creating narrow passages
(Sec.~\ref{sec:4_manipulation_region}). For the rest of regions that are guided from the base regions, we sample configurations of the base body, while the manipulator has a predefined configuration.
We also suggest a region-specific k-nearest neighbor search for effectively connecting samples generated differently by these two different regions
(Sec.~\ref{sec:4_nearest_neighbor_search}).

To validate our method, we compare our method with the coupled and decoupled
approaches for the base and manipulator in six different problems with
different characteristics (Sec.~\ref{sec:test_scenes}).  
Overall, we are able to observe that our method works robustly across
different scenes, while the other tested methods work well in some cases, but not in other cases (Sec.~\ref{sec:results_overall}).
Furthermore, we test our approach with the mobile Hubo in a real environment
mimicking a convenient store and observe similar results (Sec.~\ref{sec:real_robot_test}).

\section{RELATED WORK} 
\label{sec:2}
In this section, we discuss prior work on planning for mobile manipulators.  

\subsection{Sampling-based planning} 
\label{sec:2_sampling_based_planning}

The sampling-based approach~\cite{Kavraki96, LK98} is one of the most prominent
strategies, which has successfully tackled down the high-dimensional problems
thanks to its scalable space representation with the probabilistic
completeness. In addition, to get efficient movement, many manipulation approaches 
have used to optimal motion planning algorithms such as Informed RRT$^*$~\cite{gammell2014informed},
BIT$^*$~\cite{gammell2015batch}, FMT$^*$~\cite{janson2015fast}, and
LazyPRM$^*$~\cite{hauser2015lazy}, since Karaman et
al.~\cite{karaman2011sampling} introduced RRT$^*$ and PRM$^*$.
Burget et al.~\cite{burget2016bi} suggested BI$^2$RRT$^*$ that extends the
Informed RRT$^*$ towards a bidirectional search for a task-constrained mobile
manipulator.  Schmitt et al.~\cite{schmitt2017optimal} proposed an optimal
manipulation planner for continuous grasps, placements, and actions by
connecting the configurations.
For achieving the almost-sure asymptotic optimality of mobile manipulation
planning, our method is based on LazyPRM$^*$, while its sampling is guided by
our proposed harmonious sampling.

Guided or biased sampling with various heuristic has been widely used and
improved the performance of the optimal planners~\cite{burns2005toward,
akgun2011sampling, nasir2013rrt, kiesel2017effort}.  While these methods guided
sampling for a given joint C-space, our harmonious sampling
adaptively adjusts the sampling dimensionality for the base and manipulator
configurations in mobile manipulation.
Interestingly, Gochev et al.~\cite{gochev2012planning, gochev2013incremental}
suggested a dimensionality-reduction algorithm to efficiently handle
high-dimensional problems with adaptive dimensionality, which is, however,
applicable to only grid-based search algorithms with resolution-completeness.
On the other hand, our method enables the adaptive dimensionality for sampling
to the sampling-based approach for designing more effective exploration to the
mobile manipulation problem.

\subsection{Mobile manipulation planning} 
\label{sec:2_mobile_manipulation_planning}

In order to plan an optimal trajectory for a mobile manipulation planning
problem, we have to consider the degrees-of-freedom (DoFs) of both robotic
manipulator and its base body. 
Unfortunately, constructing a random geometric graph in such high-dimensional
C-spaces with conventional sampling-based planners can be a
significant burden due to the curse of dimensionality.

To alleviate the complexity, a divide and conquer strategy has been widely
used~\cite{chitta2012perception, srivastava2014combined, garrett2017sample}.
By partitioning the entire planning problem into a set of sub-problems (e.g.,
goal configuration generation and path planning separately for base body and
manipulator), it can efficiently subdivide the search space into a set of sub-problem,
while achieving a reasonable quality of the robot trajectory. 

For goal configuration generation, there have been several works using
reachability inversion~\cite{vahrenkamp2013robot, dong2015orientation, hertle2017identifying}. 
These approaches compute a set of suitable base positions reaching the target end-effector pose in consideration of reachability for the manipulator.
The goal configurations can be then computed by inverse kinematics (IK) with
the constructed reachability map to identify better robot placements.

Given a target end-effector pose in the workspace, various goal configurations
can be found during the planning process by adding and considering multiple
goal
configurations~\cite{stilman2007manipulation, berenson2009goalregions}.  
We adopt this approach of considering multiple goal configurations 
to find better ones among possible goal configurations.  Since our harmonious
sampling effectively explores the joint C-space, we can naturally
consider these multiple goal configurations for the base and the manipulator 
in the planning process.

\section{Overview}
\label{sec:3}

In this section, we give an overview of our approach.  We present our adaptive
sampling algorithm to effectively explore the whole configurations of a given
mobile manipulation problem.

The main task of mobile manipulation is to reach a given target end-effector
pose, $p_{goal}$, in the task space. Due to the high dimensionality of the
mobile manipulation problem, separate planning, i.e., decoupled planning,  of the base body and
manipulator is
commonly adopted.
Unfortunately, the manipulator not only has high DoFs, but also is easily
placed in a narrow space, because the configurations computed from the given
target pose
$p_{goal}$ are typically located very close to  obstacles such as shelves
and objects.  Manipulation in this narrow space with the decoupled planning may
lead to sub-optimality and increase the overall planning time due to the fixed
base chosen from the separately-run base planning for manipulation planning.  
To
alleviate these problems, we explore the whole C-space effectively
through our harmonious sampler guided by a low-dimensional base space. 

In this paper, we assume an omnidirectional mobile manipulator robot, whose
base configuration can be represented by three parameters: 2D position and its
orientation ($q_x, q_y, q_{\theta}$).  We define its space to be a base space,
$R$, and partition the 3D base space $R$ into a set of manipulation regions
$R_m$ and the rest $R_b (= R \setminus R_m)$.  We represent these regions in a
grid form for efficient processing.
Intuitively, $R_{m}$ corresponds to difficult regions such that simultaneously
adjusting the base, $b$, and manipulator, $m$, is required.  As a result, in
these manipulation regions, high-dimensional planning in the joint C-space is necessary to get a
feasible solution.
On the other hand, in the base regions $R_b$, mainly considering the DoFs of the
base body is performed to reduce the dimensionality in less cluttered regions.
Note that our sampling process is guided by the 3D base space $R$, but all of
generated samples reside in the joint C-space consisting of DoFs of both the
base and manipulator; the samples generated with the base regions $R_b$ have a
predefined configuration for joints.

\setlength{\textfloatsep}{5pt} 
\begin{algorithm}[t] 
	\vspace{0.1cm}
	\DontPrintSemicolon 
	\KwIn { $p_{goal}$: target end-effector pose, $q_{pre}^m$: predefined configuration of joints. $q_{init}$: initial configuration}
	$V \leftarrow \{q_{init}\}, E \leftarrow \emptyset, Q_{goal} \leftarrow \emptyset $\;
	\HiLi $R \gets IdentifyManipulationRegions(p_{goal})$\;
	\HiLi $\mathnormal{H} \gets SetHarmoniousSampler(R)$\;
	\While{$Termination \ condition \ is \ not \ satisfied $}{
		\If{$Q_{goal} = \emptyset$ or $rand(0, 1) < \rho_{goal}$} { 
			\HiLi $q \gets AddGoalConfiguration(p_{goal})$\;
			\If {$IsCollisionFree(q)$}{
				Insert $q$ to $V$ and $Q_{goal}$ \;
			}
			\Else{
				continue \;
			}
		}
		\Else { 
			\HiLi  $q \gets HarmoniousSampling(\mathnormal{H}, q_{pre}^m)$\;
			\If {$IsCollisionFree(q)$} {
				Insert $q$ to $V$ \;
			}
			\Else{
				continue \;
			}
		}
		\HiLi $Q_{near} \leftarrow RegionSpecificNearestNeighbor(q, R)$\; 
		\ForEach{$q_{near} \in Q_{near}$} {
			Insert $(q_{near}, q)$ to $E$ \; 
		}
		UpdateSolutionPath($G, Q_{goal}$) \; 
	}	
	
	\Return{$SolutionPath(G)$}\;
	
	\caption{The proposed algorithm} \label{alg:main_algorithm}
\end{algorithm}

Alg.~\ref{alg:main_algorithm} shows our overall approach based on
LazyPRM$^*$~\cite{hauser2015lazy}.
We briefly explain how LazyPRM$^*$ constructs a graph $G=(V,E)$ consisting of a
vertex set, $V$, and a edge set, $E$, with our proposed methods 
highlighted in the yellow color. 
We first identify $R_m$ using the reachability map and
generalized Voronoi graph (GVG), and then construct a sampling distributor to
harmonize sampling of two identified regions, $R_m$ and $R_b$ (line 2-3 of
Alg.~\ref{alg:main_algorithm}). 
We design such a sampler, a harmonious sampler $\mathnormal{H}$, to consider
the ratio of hyper-volumes of these two different sampling spaces to draw more
samples on a larger space (line 12 of Alg.~\ref{alg:main_algorithm}).  

By drawing more samples in a larger space,
there exists a connection issue between samples generated differently from
those two different regions.
To ameliorate this issue, we propose a region-specific k-nearest neighbor
(k-NN) search that effectively links between samples of different regions (line
17 of Alg.~\ref{alg:main_algorithm}).
In addition, to select the best one among various goal configurations, $Q_{goal}$, from a target
end-effector pose $p_{goal}$, we dynamically add 
goal configurations in the planning process (line 6 of
Alg.~\ref{alg:main_algorithm}).  Lastly, we compute the shortest path from
$q_{init}$ to $q_{goal} \in Q_{goal}$ on the constructed graph $G$, and then
check collisions for edges of the path (line 20 of
Alg.~\ref{alg:main_algorithm}).  This is because LazyPRM$^*$ delays edge
collision checking after finding a better path to reduce the overhead of
collision checking (line 18-19 of Alg.~\ref{alg:main_algorithm}).

\section{Main approach}
\label{sec:4}

\begin{figure}[t]
	\vspace{0.1cm}
	\centering 
	\subfigure [Reachability map.] {
		\includegraphics[width=1.57in]{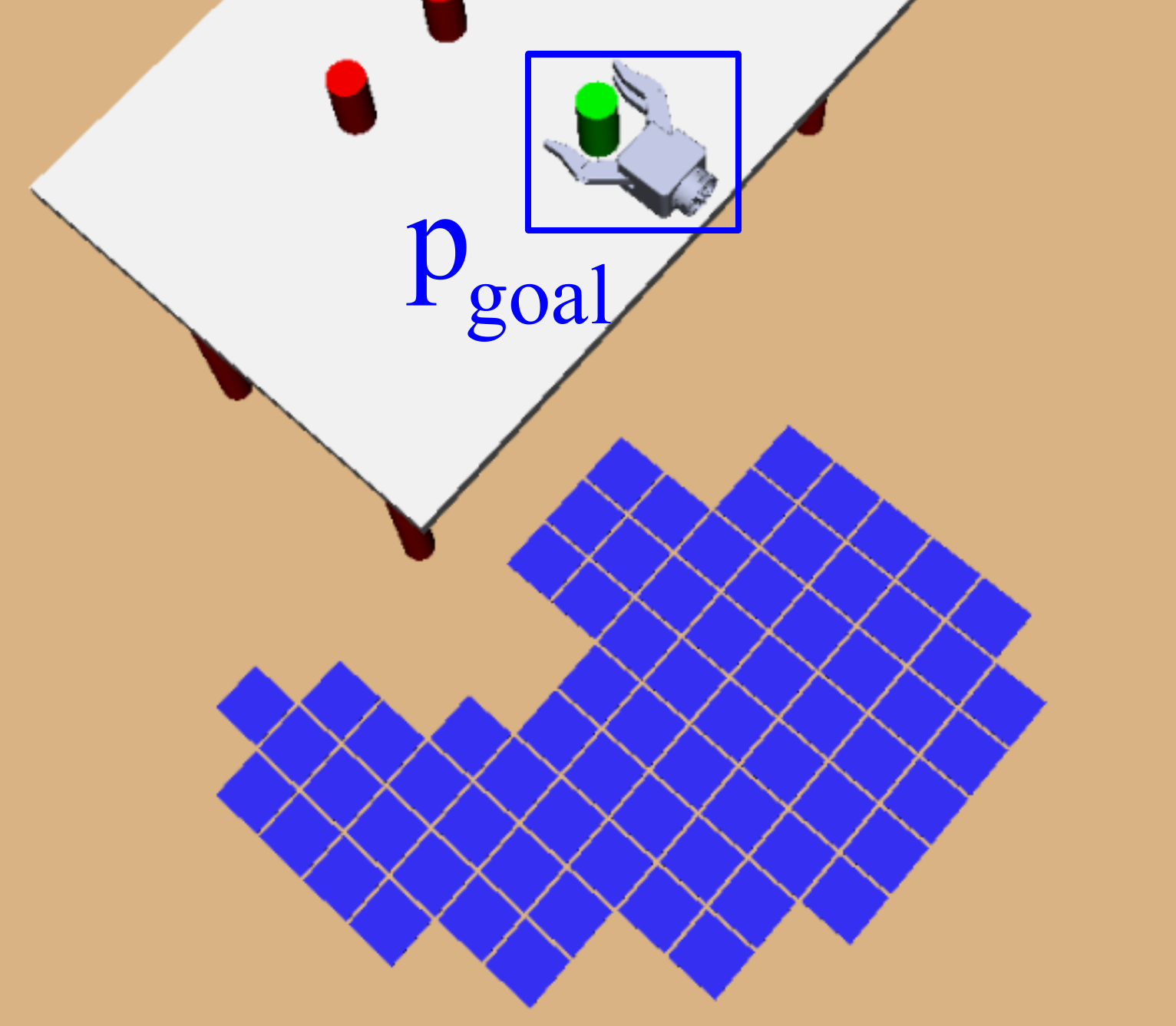}
		\label{fig:manipulation_region} 
	} 
	\subfigure [A narrow passage.] { 
		\includegraphics[width=1.568in]{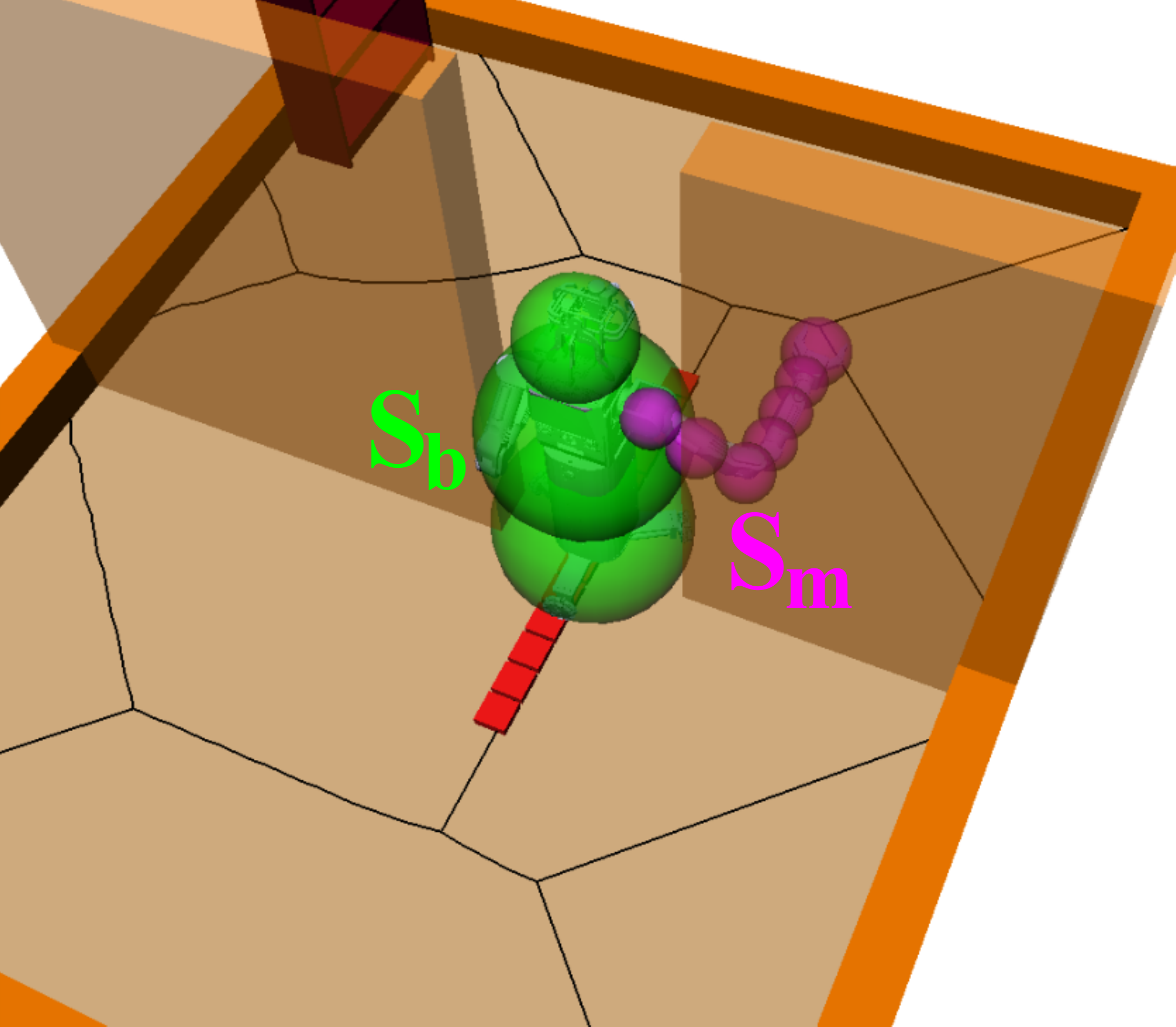}
		\label{fig:narrow_space} 
	} 
	\caption{
		These figures show our manipulation regions $R_{m}$ (visualized
		in 2D) computed by the target end-effector pose
		$p_{goal}$ and the narrow passage. $(a)$ The blue
		cells indicate reachability maps computed from the target
		end-effector pose (blue box).  $(b)$ The green and purple spheres
		represent the geometric volumes of the base $S_{b}$ and the
		manipulator $S_{m}$, respectively.  
		The red cells on the 2D floor
		visualize manipulation regions caused by obstacles such
		that $S_b$ has no collision, but $S_m$ has collision.
		Black lines on the floor show the edges of the GVG.} 
	\label{fig:approach}
	\vspace{0.0cm}
\end{figure}

In this section, we explain how to identify manipulation regions and how to
sample them harmoniously across different regions.  
We also introduce a region-specific k-nearest neighbor (k-NN) search 
for effective planning with our harmonious sampling.

\subsection{Manipulation Region Identification} 
\label{sec:4_manipulation_region}

We compute the manipulation regions $R_{m}$ in the base space, to efficiently
identify difficult regions in the joint C-space where simultaneously adjusting
the base and the manipulator.
To identify such regions,
we adopt the reachability map~\cite{vahrenkamp2013robot}
from the target end-effector pose $p_{goal}$, and utilize the GVG 
to deal with narrow passage problems caused by various objects.
Note that while we compute the manipulation regions in the base space, their sampling is done in the joint C-space consisting of the base and manipulator.

\noindent
\textbf {Reaching the target end-effector pose.}
The manipulator is required to move to reach the given target end-effector pose
$p_{goal}$.  For realizing the task, we first aim to locate our robot in
an appropriate base location in the base space $R$.
For computing those base locations, we use a
reachability map from the end-effector pose.

The reachability map defined in the 3D base space represents a set of suitable
base configurations computed for realizing the target end-effector pose
$p_{goal}$.
The blue cells in Fig.~\ref{fig:manipulation_region} are 2D example cells of the
reachability map computed from the target end-effector pose shown in the
figure.  When the base is located in the reachability map, it is guaranteed to
reach the target end-effector pose~\cite{vahrenkamp2013robot}.  We set our 3D manipulation
regions to cover those reachability map.
Once a random sample for the base configuration is generated in the manipulation region, 
we also sample a random configuration for the manipulator, to explore the
whole high-dimensional space including the base and manipulator configurations.
Details on sample generation is in Sec.~\ref{sec:4_harmonious_sampling}.

\noindent
\textbf {Avoiding obstacles in narrow passage.}
When we have many obstacles, it can potentially create narrow passages.  In
these narrow passages, it is required to simultaneously adjust the base and
manipulator to identify a collision-free path.  As a result, we expand the
manipulation regions $R_{m}$ to include such narrow passages as well.
 
To effectively identify such narrow passages, 
we construct GVG for the workspace analysis.
Each vertex and edge of GVG is associated with its closest obstacles, and thus accessing
those edges provides useful information on identifying narrow
passages~\cite{kim2014cloud}.
We also adopt a \emph{sphere expansion}
algorithm~\cite{brock2005efficient}, by simplifying the geometric description
of a robot as a set of spheres. 
Specifically, we first represent the geometric volume of the base and
manipulator by spheres, which are represented by $S_b$ and $S_m$, respectively.
We traverse edges of GVG in 2D workspace ($x$, $y$) and perform collision detection between our sphere representations with a random orientation and the environment.  We then identify regions where $S_b$ has
no collision, but $S_m$ has collision to be difficult regions for manipulation,
and thus add them to our 3D manipulation regions $R_m$.
Fig.~\ref{fig:narrow_space} shows an example of GVG in the tested environment and our sphere representations of the base and the manipulator of the predefined pose.

\setlength{\textfloatsep}{5pt} 
\begin{algorithm}[t] 
	\vspace{0.1cm}
	\DontPrintSemicolon 
	\KwIn { $\mathnormal{H}$: harmonious sampler, $q_{pre}^m$: predefined configuration of joints.} 
	\If{$rand(0, 1) < \rho_{sample}$} {
		$q_{rand}^{b} \gets GetRandomSampleForBase(\mathnormal{H})$\;
		\If{$InManipulationRegions(q_{rand}^{b})$} { 
			$q_{rand}^{m} \gets GetRandomSampleForManipulator()$\;
		} 
		\Else { 
			$q_{rand}^{m} \gets q_{pre}^{m}$\; 
		} 
	}
	\Else {
		$q_{rand} \gets UniformSampler()$\;
	}
	\Return{$q_{rand}$}\;
		
	\caption{Harmonious sampling} \label{alg:harmonious_sampler}
\end{algorithm}

\subsection{Harmonious Sampling} 
\label{sec:4_harmonious_sampling}

The purpose of our harmonious sampler $\mathnormal{H}$ is to balance the
sampling distribution between two different regions, i.e., $R_{m}$ and
$R_{b}$. 
To achieve our goal, we design $\mathnormal{H}$ to adjust the sampling
probability adaptively by considering the ratio of hyper-volumes of these two
different sampling spaces to draw more samples on a larger space.  Note that
the manipulation regions $R_m$ is defined in the base space $R$, but is
associated with the larger sampling space consisting of the base and
manipulator, i.e., the joint high-dimensional C-space.

Our harmonious sampler $\mathnormal{H}$ first generates a random sample for the
base configuration $q_{rand}^{b}$ (line 2 of Alg.~\ref{alg:harmonious_sampler}) and
then we check whether the sample is in either $R_{m}$ or $R_{b}$.  If
$q_{rand}^{b} \in R_{m}$, we also sample a random configuration of the
manipulator, $q_{rand}^{m}$ (line 3-4 of Alg.~\ref{alg:harmonious_sampler}). Otherwise, $q_{rand}^{m}$ is initialized with 
predefined values of joints, $q_{pre}^{m}$, which can be initial values 
of joints (line 5-6 of Alg.~\ref{alg:harmonious_sampler}).
Note that $q_{rand}$ is composed of $q_{rand}^{b}$ and $q_{rand}^{m}$(=$q_{rand} \setminus q_{rand}^{b}$).

To generate a random sample for the base configuration $q_{rand}^{b}$
according to our harmonious sampler $\mathnormal{H}$, we use a probability
mass function (PMF) defined in a grid form
over the 3D base space. Each cell of the grid over the 3D base space has only
two types of indicating whether the cell is in $R_m$ or $R_b$. 
Our harmonious sampler $\mathnormal{H}$ then uses the inverse transform
sampling method that works for an arbitrary sampling PMF. 

To compute a sampling probability for two regions $R_m$ and $R_b$, we utilize
the notion of the sampling hyper-volume associated to each cell, $\gamma_c$.
Specifically, when a cell  $\gamma_c$
is on the manipulation region $R_m$, its
sampling probability is computed based on the hyper-volume of its sampling space, $\nu(\gamma_c)$, which
is simply calculated by multiplying the intervals between the
lower bound, $q_d^{min}$, and upper bound, $q_d^{max}$, for each parameter
space $d$ of its sampling space:  
\begin{equation} 
\nu (\gamma_{c} | \gamma_c \in R_m) = \prod_{d=1}^{D(R_m)} w_{d}*(q_{d}^{max} - q_{d}^{min}),
\label{eq:density}
\end{equation}
where $D(R_m)$ is the DoF associated with the sampling space of $R_m$,
and $w_d$ is the weight to match the units for each dimension (Sec.~\ref{sec:4_distance_metric}).
This approach is also used for computing a
sampling probability for cells on the base regions $R_b$.

The PMF for our harmonious sampler $\mathnormal{H}$ of cells 
$\gamma_c$ in the base space $R$ can be
defined as:
\begin{equation} 
\mathbb{P}(\gamma_c | R) = \nu(\gamma_c) / \sum\limits_{i=1}^N \nu(\gamma_i),
\label{eq:probability_density} 
\end{equation}
where $N$ is the number of cells in the base space $R$, whose cell index is $i$.
Consequently, the harmonious sampler $\mathnormal{H}$ varies the sampling
probability according to sampling DoFs associated with each cell, i.e. three sampling DoFs for $R_b$
and ten sampling DoFs for $R_m$.

Note that our harmonious sampler $\mathnormal{H}$ is simple yet useful, but
cannot guarantee the optimality.
To ensure the probabilistic completeness and the asymptotic optimality of our
harmonious sampling, we use both our harmonious sampler $\mathnormal{H}$
at a probability of $\rho_{sample}$ and the uniform sampler 
over the entire sampling space at a probability of
$(1-\rho_{sample})$~\cite{akgun2011sampling, nasir2013rrt}
(Alg.~\ref{alg:harmonious_sampler}).

\subsection{Linking Different Regions} 
\label{sec:4_nearest_neighbor_search}

\begin{wrapfigure}{r}{4cm}
	\centering
	\includegraphics[width=1.55in]{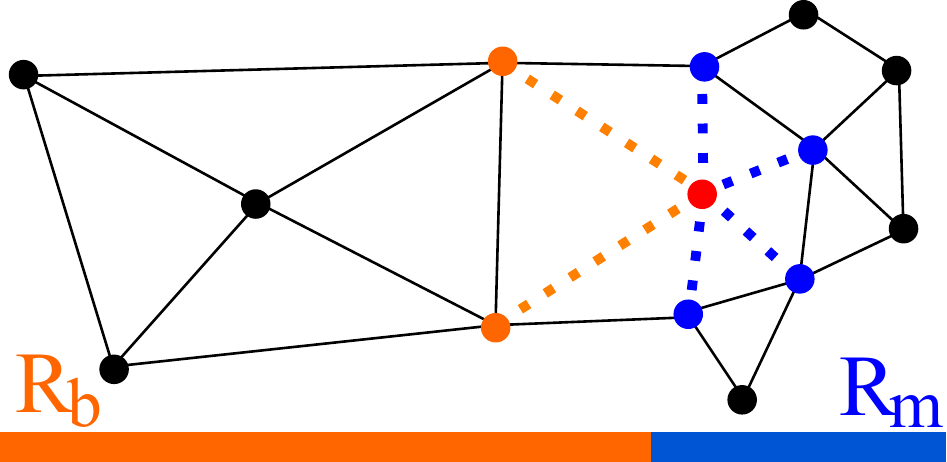}
	\caption{
		Our region-specific k-NN links samples generated through
		different regions. Blue and orange dotted lines, i.e., connections,  are made
		by our approach for the red dot.
	} 
	\label{fig:region_specific_nn}
\end{wrapfigure}

Our harmonious sampling adaptively adjusts its sampling space depending on
whether a sample is generated on the base regions $R_b$ or manipulation regions
$R_m$.
Note that we generate more samples on the joint C-space associated with $R_m$,
compared to that associated with $R_b$, thanks to the wider sampling DoF
associated with $R_m$.
Fig.~\ref{fig:region_specific_nn} illustrates the different sampling densities
on the joint C-space; in this example, $R_m$ and $R_b$ are illustrated in the
bottom one-dimensional space. 

We found that this varying sampling probability causes a sub-optimal performance,
especially due to sub-optimal linking along the boundary of two different
regions.
For example, when we naively perform k-NN search for samples (e.g., the red dot in Fig.~\ref{fig:region_specific_nn}) generated from the
manipulation regions $R_m$, identified samples are likely to be ones from $R_m$
due to its high sample probability causing small distances between samples
generated from  $R_m$
(Fig~\ref{fig:region_specific_nn}).

To alleviate this connectivity issue between two different regions, we propose
to use a simple region-specific k-NN search.  For samples associated with the
manipulation regions, we perform two independent k-NN search: one is performed
only with samples associated with $R_m$, resulting in blue dotted lines in
Fig.~\ref{fig:region_specific_nn}, and the other k-NN is performed only with
samples associated with $R_b$, resulting in the orange dotted lines in
Fig.~\ref{fig:region_specific_nn},   for improving the connectivity between two
different regions.  For samples
associated with the base regions, we simply use the original k-NN search that is
performed with all the available samples.

We found that our region-specific k-NN search may generate unnecessary edges
between samples that are located in different regions, causing a more memory
overhead, approximately 25\% over using the original k-NN search.
Fortunately, our method is based on LazyPRM$^*$~\cite{hauser2015lazy}, 
and thus there is not much difference in terms of the number of edge collision
checking, since we perform edge collision checking on the solution paths.
As a result, using region-specific k-NN search improves the path costs by 7\%
in our tested environments without a significant runtime performance overhead,
thanks to better connectivity between two different
regions (Table~\ref{tab:result_analysis}).

\subsection{Distance Metric} 
\label{sec:4_distance_metric}

We perform planning in the joint C-space consisting of the base and
manipulator.  Since the base and joints have
different kinds of quantities, i.e., linear and angular quantities, we have the
incompatibility issue, which can lower the efficacy of performing nearest
neighbor search during planning with respect to the asymptotic
optimality~\cite{karaman2011sampling}.

In case of the incompatibility, LaValle et al.~\cite{lavalle2006planning}
considered a robot displacement by using coefficients of different
quantities as weights, $w$, to match the units, as the following:
\begin{equation}
dist = \sqrt{\sum_{d}^{D} (w_d * \rVert q_d-q'_d \rVert) ^2},
\label{eq:distance_function}
\end{equation}
where $D$ is the DoF of the joint C-space; for the mobile Hubo robot,
$q_x, q_y, q_{\theta}$ correspond to the base configuration, and the rest to the
manipulator.

We explain how to determine weights $w_i$ to match the units using the robot
displacement.  For two configurations $q$ and $q'$, the robot displacement metric,
$disp (\cdot, \cdot)$, can
be defined as: 
\begin{equation}
disp(q, q') = \max_{a \in \mathnormal{A}} {\rVert \tau(q, a)- \tau(q', a) \rVert},
\label{eq:displacement_metric}
\end{equation}
where $\tau (q, a)$ is the position of point $a$ for the robot $\mathnormal{A}$
with the configuration $q$ in the workspace.

Since the robot displacement metric $disp(\cdot,\cdot)$ yields the maximum
amount in the workspace,
the first joint with the largest change of the manipulator should
have a larger weight.  
While fixing weights of $q_x, q_y$ to $1$ for the simplicity, the maximum
displacement for each angular quantity is treated as arc lengths in the
workspace that can be made given its joint. 
One can then show that the weight for a joint of a manipulator can be computed
by the distance from the joint to the end-effector; the weight intuitively corresponds to the
maximum arc length per radian.

\noindent
\paragraph{Supporting multi-goal configurations.} 
Given the target end-effector pose $p_{goal}$,
our work supports multi-goal configurations in the joint 
C-space by planning the base and manipulator simultaneously with our harmonious
sampling.  Since possible goal configurations from $p_{goal}$ can be infinite,
our method dynamically adds goal configurations by injecting the reachability map and IK
solver~\cite{berenson2009goalregions}
during planning according to a
probability of $\rho_{goal}$ (line 5-6 of Alg.~\ref{alg:main_algorithm}).  Our
harmonious sampler $\mathnormal{H}$ then generate samples in the joint C-space
according to $R_{m}$, naturally supporting dynamically added goal configurations.

\section{Experiments and Analysis}
\label{sec:5}
\begin{figure}[t]
	\vspace{0.3cm}
	\centering
	\subfigure [Grasping an object on the table (Problem 1).]
	{ 
		\includegraphics[width=1.5in]{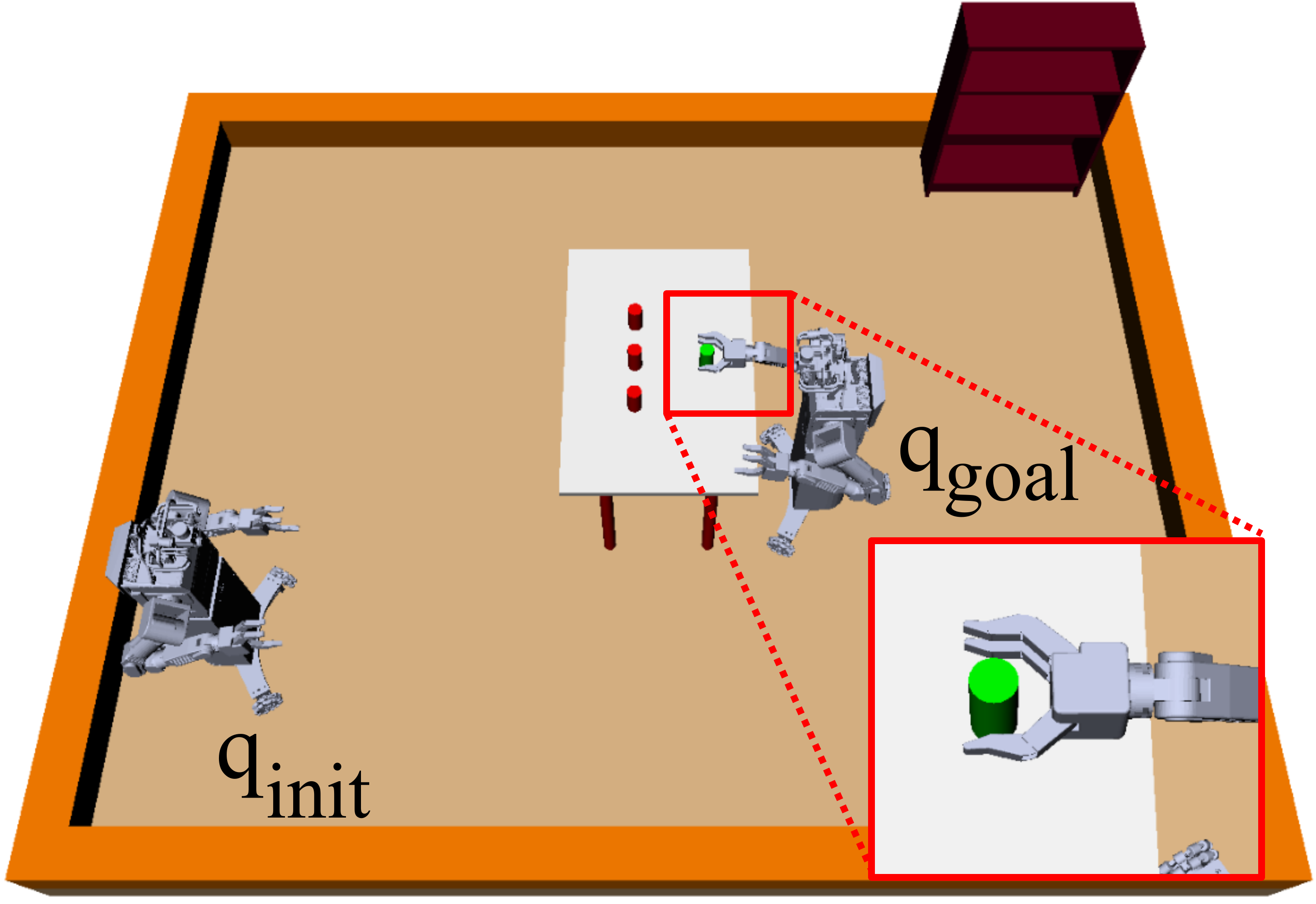}
		\label{fig:scene_1}
	}
	\hspace{0.08in}
	\subfigure [Grasping an object on the table with ground obstacles (Problem 2).]
	{ 
		\includegraphics[width=1.5in]{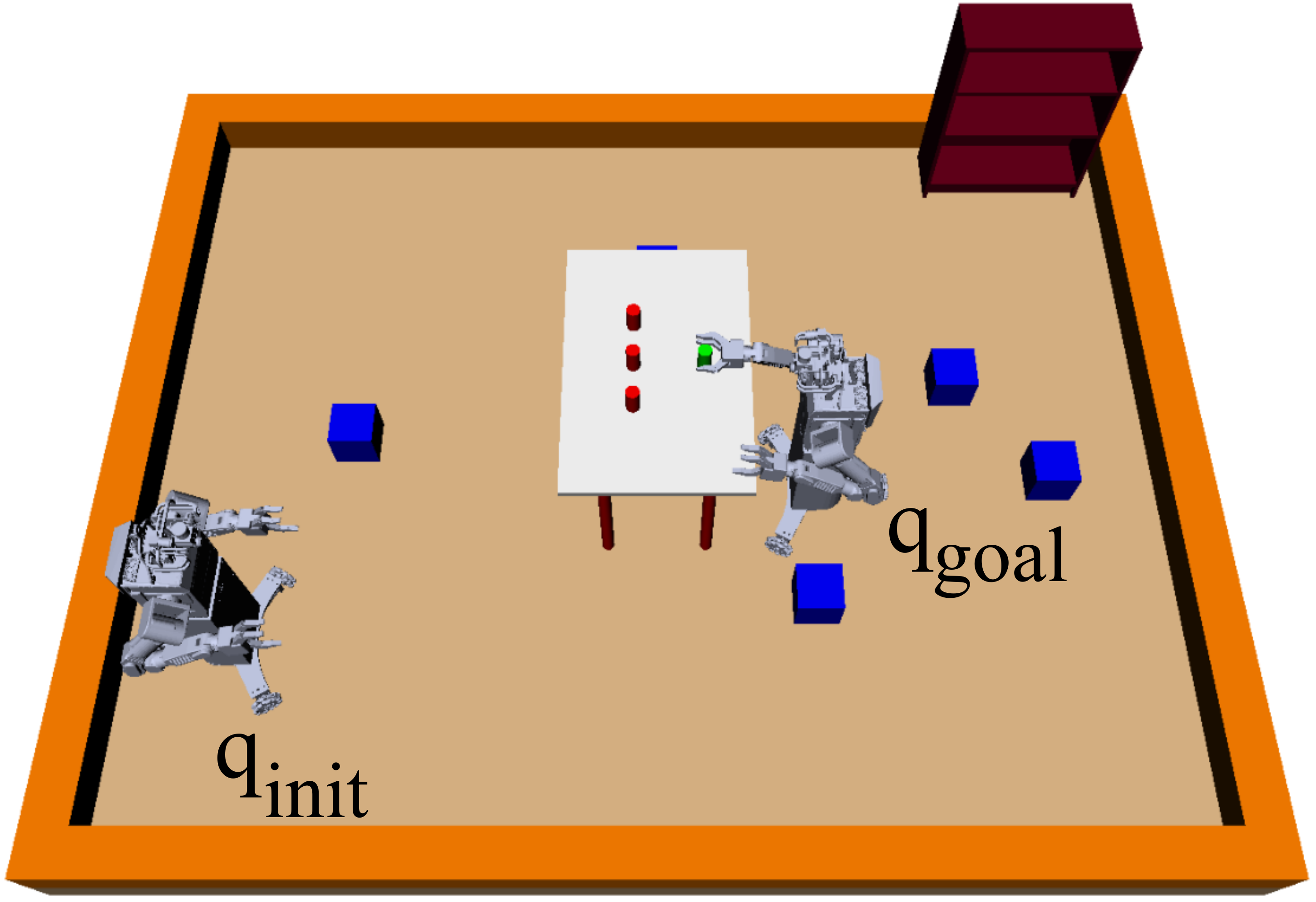}
		\label{fig:scene_2}
	}
	\subfigure [Grasping an object on the shelf (Problem 3).]
	{ 
		\includegraphics[width=1.5in]{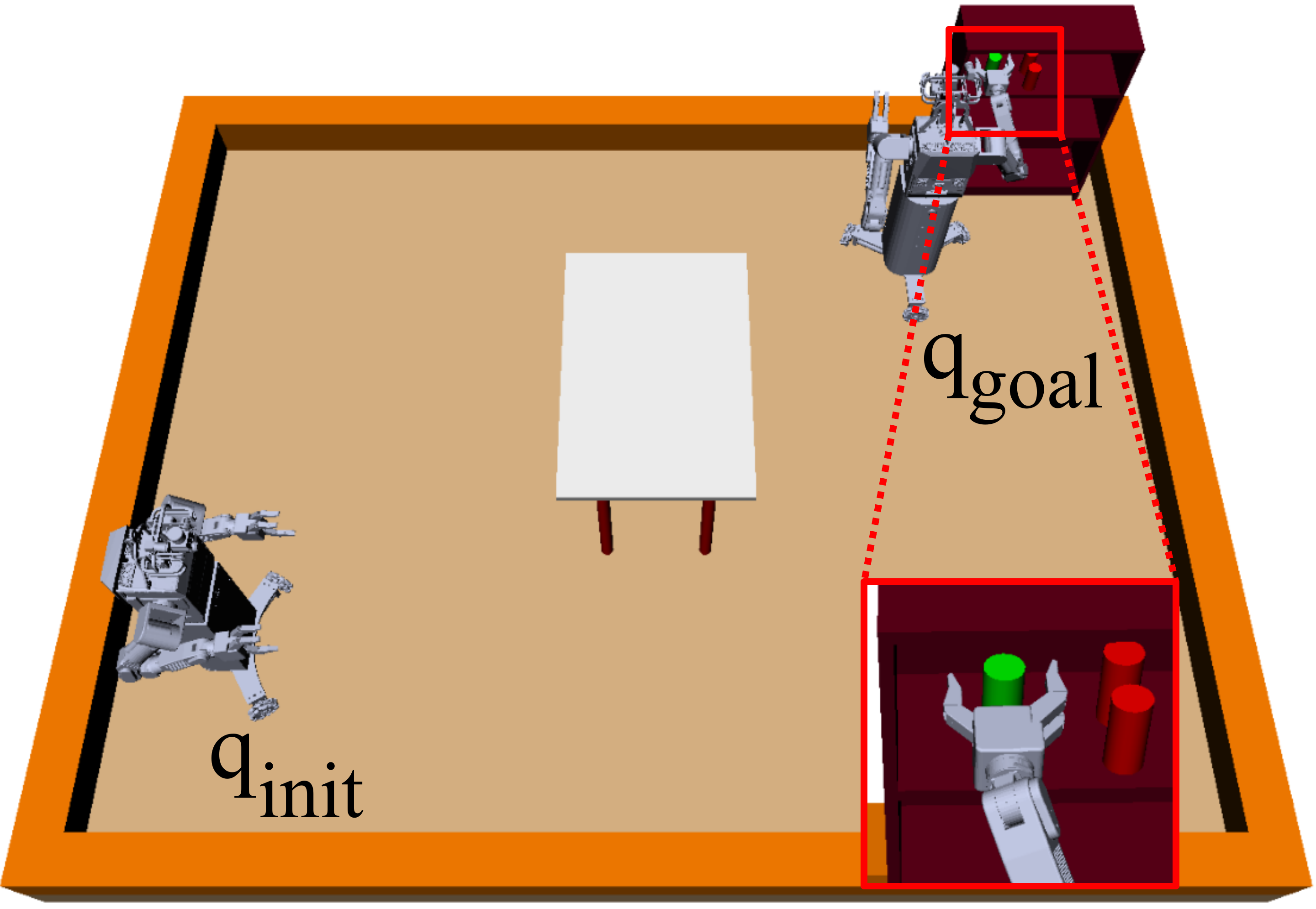}
		\label{fig:scene_3}
	}
	\hspace{0.08in}
	\subfigure [Grasping an object on the shelf with ground obstacles (Problem 4).]
	{ 
		\includegraphics[width=1.5in]{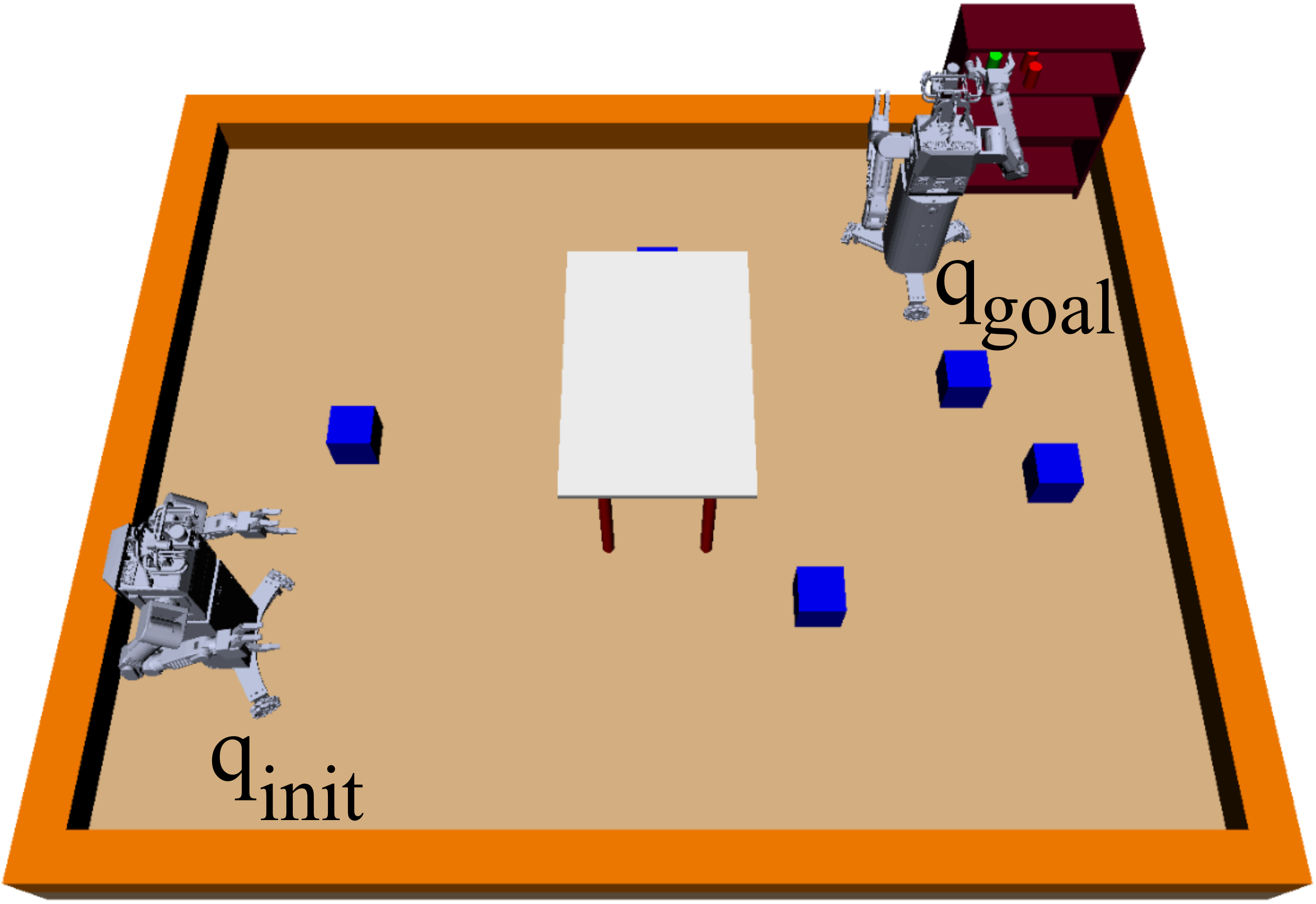}
		\label{fig:scene_4}
	}
	\subfigure [The worst case of the predefined posture (Problem 5).]
	{ 
		\includegraphics[width=1.5in]{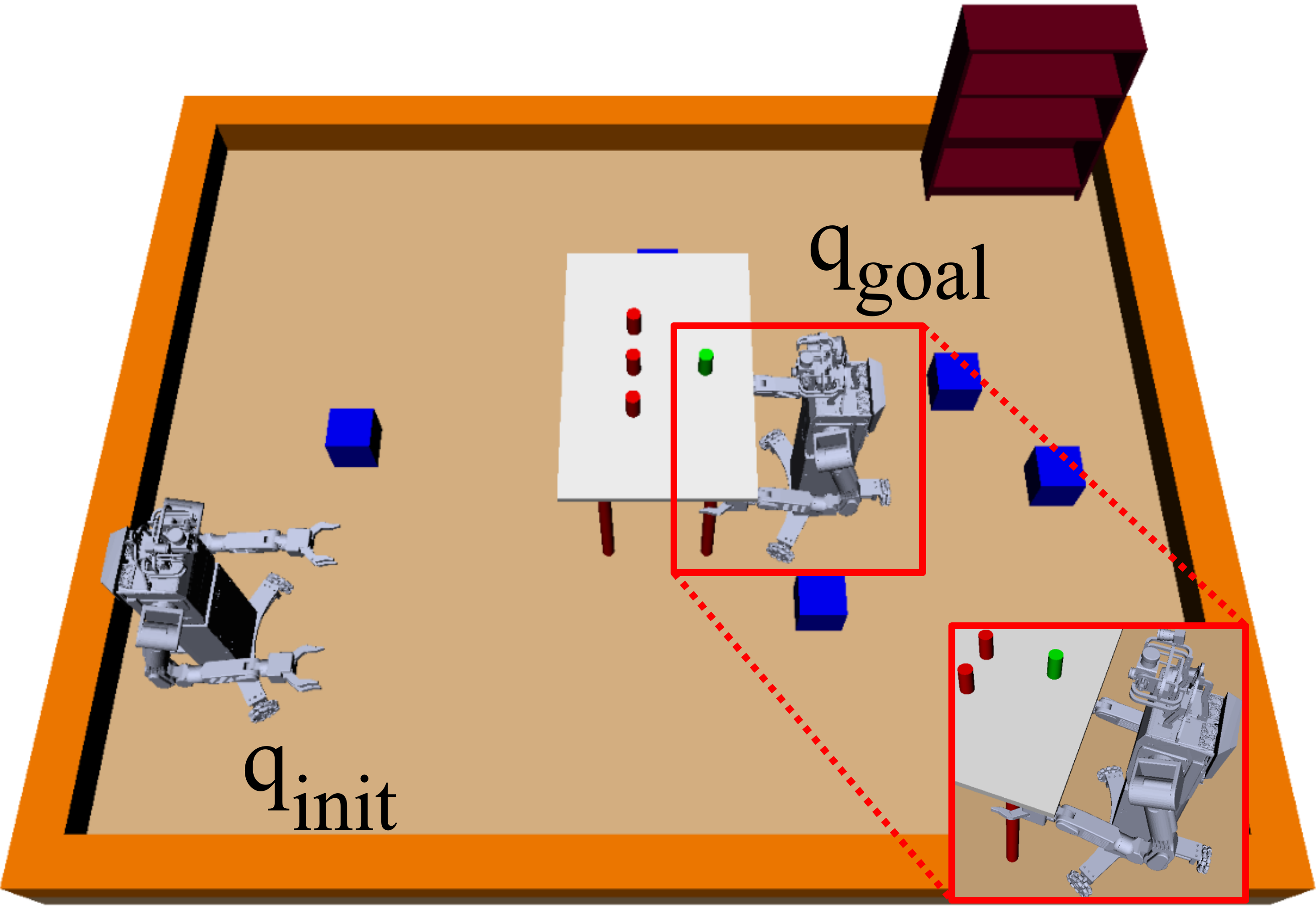}
		\label{fig:scene_5}
	}
	\hspace{0.08in}
	\subfigure [The narrow passage (Problem 6).]
	{ 
		\includegraphics[width=1.5in]{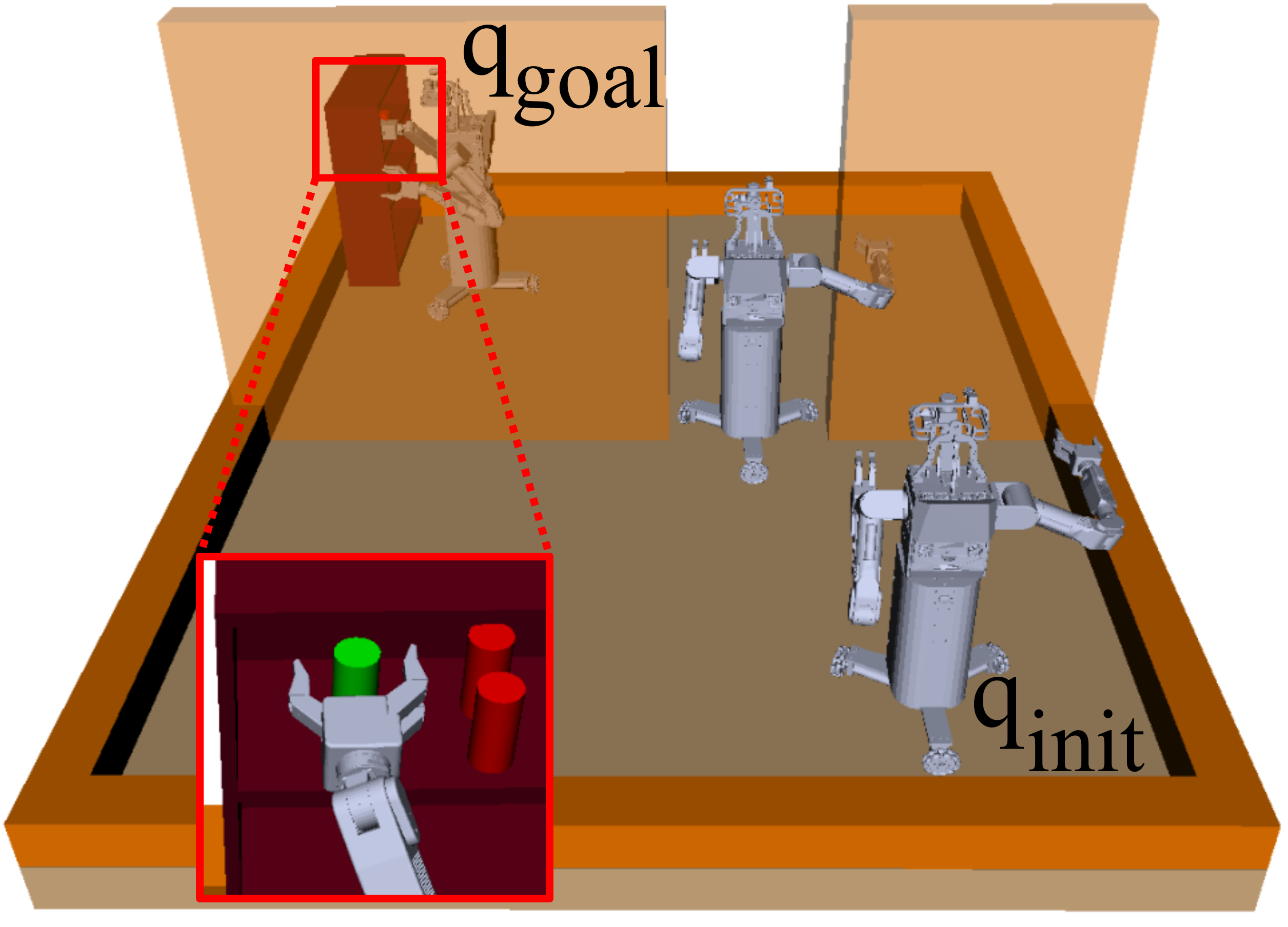}
		\label{fig:scene_6}
	}
	\caption{
		This shows our test scenes with the start and goal configurations, $q_{init}$ and $q_{goal}$. 
		$(a)$ and $(b)$ are to grasp an object on the table. $(c)$ and
		$(d)$ are to grasp an object on the shelf.  $(b)$ and
		$(d)$ consist of additional objects, obstructing
		the robot.  $(e)$ shows a difficult case of the predefined posture
		in $(b)$.  The manipulator is located under the table.
		$(f)$ shows the narrow passage where adjusting a manipulator is
		needed to move a base.
	}
	\label{fig:scene}
	\vspace{0cm}
\end{figure}

We test our method on a machine that has 3.40GHz Intel i7-6700 CPU and 16GB
RAM.  We use the simulated mobile Hubo robots for testing our method within the
OpenRAVE simulator (Fig.~\ref{fig:scene}). We also use a real mobile Hubo robot~\cite{lee2019hubo}
for testing the usability in a real environment (Fig.~\ref{fig:main}).
A manipulator of the robot has seven joints, and its base consists of $x$, $y$ and $\theta$.
We partition the base space $R$ into a grid form using $0.1m$ resolution for
$x$, $y$ and $30$ degree resolution for $\theta$.  We use
LazyPRM$^*$~\cite{hauser2015lazy} as our base optimal planner, and set $\rho_{sample}$
$0.8$ and $\rho_{goal}$ $0.01$ for our tests.

To validate our method, we compare our method against two different approaches: the decoupled and coupled approaches.  The decoupled
approach~\cite{chitta2012perception} divides the planning process into two
separate planning for the base and manipulator, while the coupled
approach~\cite{burget2016bi} performs the planning within the joint C-space
representing both base and manipulators and thus having 10 DoFs.  

\begin{figure*}[t]
	\vspace{0.1cm}
	\centering
	\subfigure [Result of Problem 3 w/ different methods]
	{ 
		\includegraphics[width=2.23in]{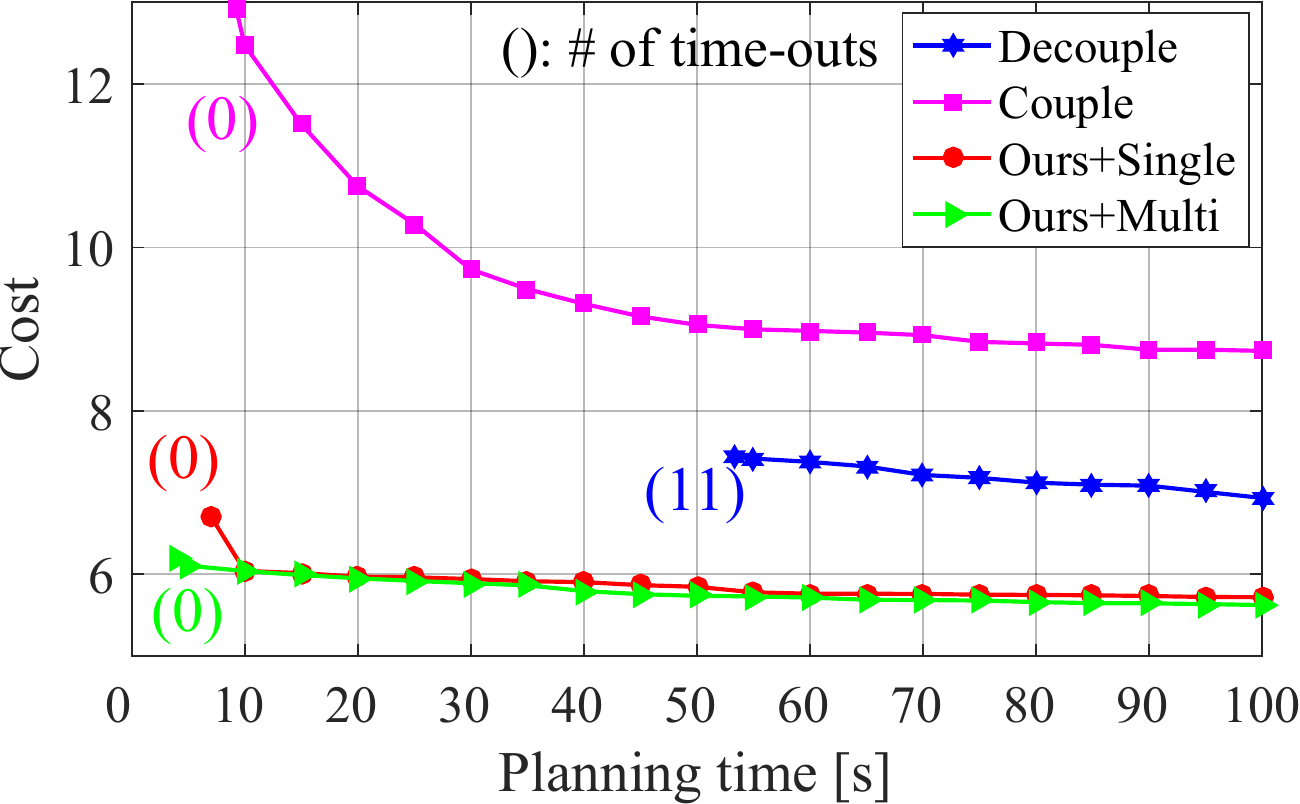}
		\label{fig:result_pro3}
	}
	\subfigure [Result of Problem 4 w/ different methods]
	{ 
		\includegraphics[width=2.21in]{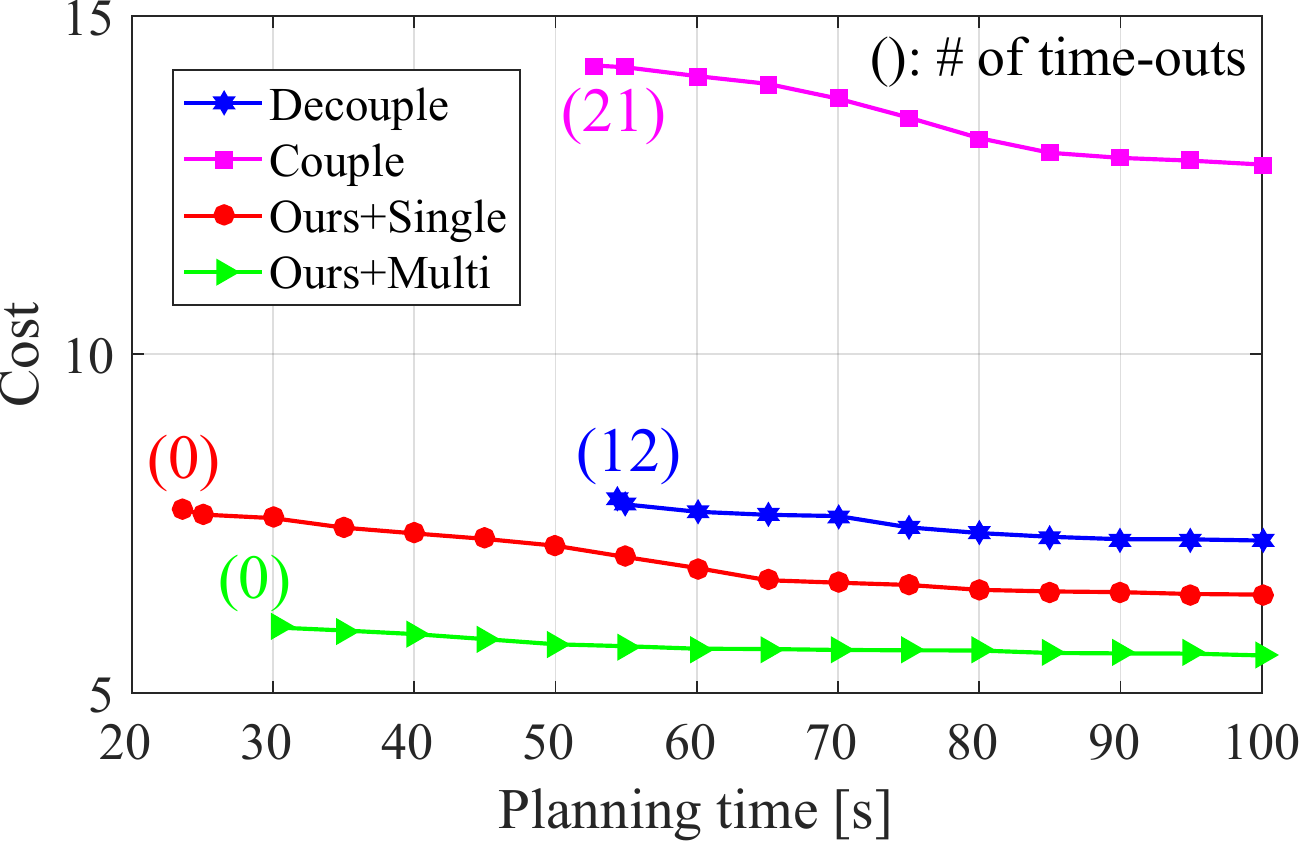}
		\label{fig:result_pro5}
	}
	\subfigure [Results w/ different time budget ratios for the decoupled
	approach in Problem 4]
	{ 
		\includegraphics[width=2.18in]{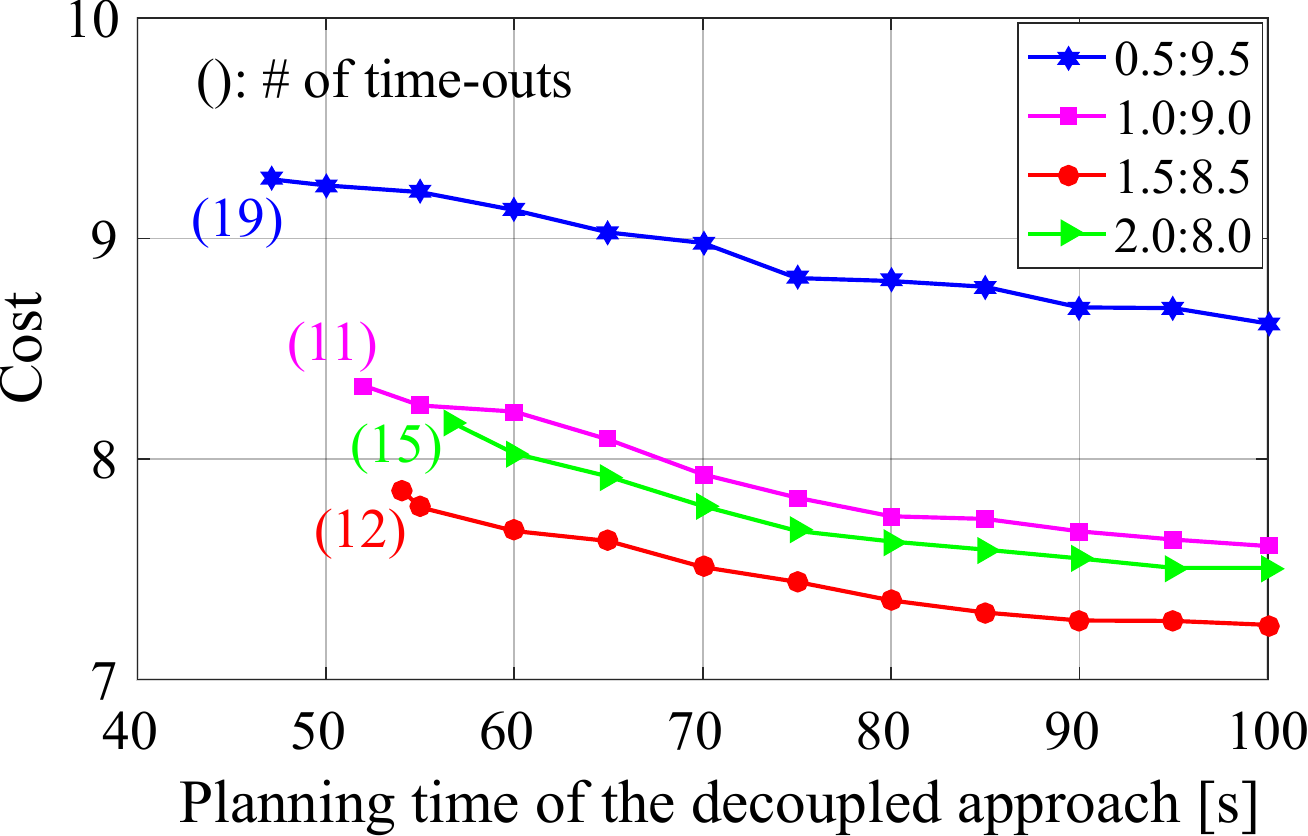}
		\label{fig:result_sep}
	}

	\caption{
		This figure shows the cost over planning time of different
		methods.
		$(a)$ and $(b)$ show that our method (Ours + Multi) achieves
		the best performance.
		$(c)$ shows that $1.5:8.5$ time budget ratio for the base and the
		manipulator used in the decoupled approach shows the best performance
		among the tested budget ratios.We visualize graphs from the time that an initial solution is computed.
	}
	\label{fig:chart_time}
	\vspace{-0.22cm}
\end{figure*}

For the decoupled planning, we need to allocate separate time budget for base
and manipulator planning parts for the same-time comparison against other
approaches. Unfortunately, this issue was not explicitly discussed in prior
techniques, and left for user setting.
We test different time budgets for the decoupled method, and found $1.5:8.5$
time budget ratio for the base and the manipulator shows the best performance
on average.  Fig.~\ref{fig:result_sep} shows results with varying ratios in one
of difficult testing environments.

We experiment for various planning with a single goal configuration
except for the multiple goal approach generating multiple goal configurations
given the target end-effector pose.  For all the experiments, we perform $50$
tests given a time budget of 100 seconds, and report their average performance.

\subsection {Test problems}
\label{sec:test_scenes}

We prepare six different problems to validate our approach, as shown in
Fig.~\ref{fig:scene}.  Our test scene is $5m \times 4m$  wide.  The first
and second problems (table scenes) are designed to grasp an object on the table
as relatively simpler problems with a few obstacles for the manipulator.  
The third and fourth problems (shelf scenes) have a narrow shelf with obstacles on the bookshelf.
Moreover, to show the effect of base movement, the second and fourth problems
additionally place ground obstacles that interfere with the movement of a robot. 

To intentionally create a difficult case that we may encounter in practice, 
we use predefined postures
for the fifth and sixth 
that are difficult for manipulating objects or passing narrow passages without
dynamically modifying
the predefined postures.
For other scenes, we use the basic posture of the mobile Hubo robot.

\begin{table}[t]
	\vspace{0.25cm}
	\centering
	\renewcommand \arraystretch{1.15}
	\setlength{\tabcolsep}{2.0pt}
	\caption{ 
		Various statistics, averaged out from 50
		independent tests. 
		For the decoupled approach, we also show separate times (s)
		spent on finding an initial solution for each of the base and
		the
		manipulator. 
	}
	\begin{tabular}{|c|c|c|c|c|c|}
		\hline
		\multicolumn{2}{|c|}{} & Decoupled & Coupled & Ours + & Ours + \\
		\multicolumn{2}{|c|}{} & Approach & Approach & Single & Multi \\
		\hline
		& \# of time-outs & 0 & 0 & 0 & 0 \\
		\cline{2-6}
		Problem 1 & Initial solution time & 5.70 & \multirow{2}{*}{11.52} & \multirow{2}{*}{\textbf{2.23}} & \multirow{2}{*}{2.45} \\
		Fig.~\ref{fig:scene_1} & (Base/Manipulator) & (0.57/5.13) & ~ & ~ & ~ \\
		\cline{2-6}
		~ & Final distance cost & 4.61 & 8.10 & 4.61 & \textbf{4.03} \\
		\cline{2-6}
		\hline
		\hline
		~ & \# of time-outs & 0 & 0 & 0 & 0 \\
		\cline{2-6}
		Problem 2 & Initial solution time & 6.55 & \multirow{2}{*}{27.72} & \multirow{2}{*}{\textbf{2.79}} & \multirow{2}{*}{3.08} \\
		Fig.~\ref{fig:scene_2} & (Base/Manipulator) & (0.93/5.62) & ~ & ~ & ~ \\
		\cline{2-6}
		~ & Final distance cost & 5.40 & 9.65 & 5.18 & \textbf{4.81} \\
		\cline{2-6}
		\hline
		\hline
		~ & \# of time-outs & 11 & 0 & 0 & 0 \\
		\cline{2-6}
		Problem 3 & Initial solution time & 39.15 & \multirow{2}{*}{9.25} & \multirow{2}{*}{7.00} & \multirow{2}{*}{\textbf{4.14}} \\
		Fig.~\ref{fig:scene_3} & (Base/Manipulator) & (0.75/38.40) & ~ & ~ & ~ \\
		\cline{2-6}
		~ & Final distance cost & 6.92 & 8.73 & 5.71 & \textbf{5.62} \\
		\cline{2-6}
		\hline
		\hline
		~ & \# of time-outs & 12 & 21 & 0 & 0 \\
		\cline{2-6}
		Problem 4 & Initial solution time & 41.50 & \multirow{2}{*}{52.74} & \multirow{2}{*}{\textbf{24.34}} & \multirow{2}{*}{30.26} \\
		Fig.~\ref{fig:scene_4} & (Base/Manipulator) & (2.05/39.45) & ~ & ~ & ~ \\
		\cline{2-6}
		~ & Final distance cost & 7.24 & 12.76 & 6.37 & \textbf{6.23} \\
		\cline{2-6}
		\hline
		\hline
		~ & \# of time-outs & 0 & 0 & 0 & 0 \\
		\cline{2-6}
		Problem 5 & Initial solution time & 9.55 & \multirow{2}{*}{31.96} & \multirow{2}{*}{3.94} & \multirow{2}{*}{\textbf{3.65}} \\
		Fig.~\ref{fig:scene_5} & (Base/Manipulator) & (2.07/7.48) & ~ & ~ & ~ \\
		\cline{2-6}
		~ & Final distance cost & 6.97 & 9.84 & 5.79 & \textbf{5.08} \\
		\cline{2-6}
		\hline
		\hline
		~ & \# of time-outs & - & 28 & 0 & 0 \\
		\cline{2-6}
		Problem 6 & Initial solution time & \multirow{2}{*}{-} & \multirow{2}{*}{70.90} & \multirow{2}{*}{36.91} & \multirow{2}{*}{\textbf{33.13}} \\
		Fig.~\ref{fig:scene_6} & (Base/Manipulator) &  & ~ & ~ & ~ \\
		\cline{2-6}
		~ & Final distance cost & - & 16.45 & 12.16 & \textbf{11.73} \\
		\cline{2-6}
		\hline
	\end{tabular}
	\vspace{0.0cm}
	\label{tab:result_overall}
\end{table}

\subsection {Result Analysis}
\label{sec:results_overall}

We compare our method with the coupled and decoupled approaches across six
different problems.
Initially, our method identifies the manipulation regions $R_m$ and
constructs  the harmonious sampler $\mathnormal{H}$ ($\approx$0.5s at the
most); these times are included in the reported planning time.
Table~\ref{tab:result_overall} shows various run-time results of the tested
methods.
Problem 1 and 2 are simple with a few obstacles, so
all the methods find a solution in reasonable planning time
(Table~\ref{tab:result_overall}).
In these simple cases, the simplest, decoupled approach finds
initial solutions faster than the coupled approach that performs sampling in
the joint
C-space.
For Problem 3, on the other hand, the coupled approach finds solutions
faster than the decoupled approach.  Furthermore, the decoupled approach did
not find solutions in $11$ of $50$ tests.
This clearly indicates the problem of the decoupled approach that
does not work well at the narrow space.
Overall, the decoupled and coupled approaches have the weakness in terms of
showing the robust performance across different scenes.

Fortunately, our methods, especially, Ours + Multi considering multiple goal
configurations, find solutions faster than
both coupled and decoupled approaches, and achieve the shortest path in all of
Problem 1 to 4.
Fig.~\ref{fig:result_pro3} and Fig.~\ref{fig:result_pro5} show how different methods behave as a function of planning time. Our method (Ours + Multi) computes shorter
paths given any fixed planning time.
These results clearly show benefits of our approach that adaptively performs
sampling for base and manipulators in their joint C-space, even in
simple and middle-level problems.

Let us now consider most difficult problems among the tested ones.
In Problem 5, our method finds the initial solution much faster, more than one
order
of magnitude for Ours + Multi, than the coupled and decoupled approaches.
In the case of Problem 6, also, ours achieves better solutions than the
coupled approach, even though the decoupled cannot solve the problem.  These
results demonstrate the efficiency and robustness of our methods.

\noindent
\textbf {Single vs. multiple goal configurations.}
Considering multiple goal configurations shows to produce smaller costs across
all the tests given the time budget (100s), thanks to considering various
candidate configurations to reach the target end-effector pose
(Table~\ref{tab:result_overall}).  Nonetheless, ours with multiple goal
configurations shows similar performance in terms of finding an initial
solution over considering only a single goal configuration.
This is because the overhead of considering multiple goal configurations is
about 2.8\% of the overall planning time, resulting in slower performance for
identifying the initial solution.  However, this overhead is small, and thus is
paid well as we have more planning time for achieving shorter paths.

\begin{table}[t]
	\centering
	\renewcommand \arraystretch{1.15}
	\setlength{\tabcolsep}{2.0pt}
	\caption{ 
		The results of the simple problem 2 and the
		difficult problem 4 with varying volume ratios for our
		harmonious sampler and region-specific NN search.
	}
	\begin{tabular}{|c|c|c|c|c|c|}
		\hline
		\multicolumn{2}{|c|}{k-NN type} & Original & \multicolumn{3}{|c|}{Region-specific} \\
		\hline
		\multicolumn{2}{|c|}{Ratio of $\nu (\gamma_{c}|\gamma_c\in R_m)$} & Ours & 0.2$\times$Ours & Ours (Eq.~\ref{eq:density}) & 5$\times$Ours \\
		\hline
		\multirow{3}{1.2cm}{\centering{Problem 2 \\ Fig.~\ref{fig:scene_2}}} & \# of time-outs & 0 & 0 & 0 & 0\\
		\cline{2-6}
		& Initial solution time & 2.99 & 5.26 & \textbf{2.79} & 3.08 \\
		\cline{2-6}
		& Final distance cost & \textbf{5.18} & 5.45 & \textbf{5.18} & 5.33 \\
		\hline
		\hline
		\multirow{3}{1.2cm}{\centering{Problem 4 \\ Fig.~\ref{fig:scene_4}}} & \# of time-outs & 4 & 1 & 0 & 1\\
		\cline{2-6}
		& Initial solution time & 31.93 & 37.19 & \textbf{24.34} & 29.61 \\
		\cline{2-6}
		& Final distance cost & 6.85 & 6.79 & \textbf{6.37} & 6.61\\
		\hline
	\end{tabular}
	\vspace{0.0cm}
	\label{tab:result_analysis}
\end{table}

\noindent
\textbf {Analysis of the harmonious sampler.}
To see benefits of our harmonious sampler $\mathnormal{H}$, we check how the performance of our
method behaves according to $\nu (\gamma_{c}|\gamma_c\in R_m)$ of
Eq.~\ref{eq:density} for Problem 2 and 4, given the region-specific NN search.
Specifically, we compare the performance of our approach (Ours) against 
decreasing and increasing $\nu (\gamma_{c}|\gamma_c\in R_m)$ by five times; they
are denoted as 0.2$\times$Ours and 5$\times$Ours, respectively.
Table~\ref{tab:result_analysis} shows that the number of time-outs, cost, and
planning time are increased when $\nu (\gamma_{c}|\gamma_c\in R_m)$ is increased or
decreased.
These results support the benefit of our harmonious sampler $\mathnormal{H}$
taking into account
the hyper-volumes of different regions.

\noindent
\textbf {Analysis of the region-specific k-NN search.}
To see the benefits of our region-specific k-NN search, we compare its
performance against 
the original k-NN search that performs k-NN search with all the samples in the
joint  C-space.
Table~\ref{tab:result_analysis} shows that in the simple problem (Problem 2), 
there is not much difference in terms of the final distance
cost between two methods.  On the other hand, in a difficult problem (Problem
4), our region-specific k-NN search (Ours) reduces all of three measures
(including the
final distance cost) over using the original k-NN search.  Specifically, using
the
region-specific k-NN search shows $1.3$ times performance improvement in
terms of finding an initial solution time, and $7\%$ cost reduction,  compared
to using the original k-NN search.

\begin{table}[t]
	\centering
	\renewcommand \arraystretch{1.15}
	\setlength{\tabcolsep}{2.0pt}
	\caption{ 
			Results with a predefined posture, which is chosen from
			five random postures.
			These results are averaged out from 50 independent
			tests. 
	}
	\begin{tabular}{|c|c|c|c|c|c|}
		\hline
		\multicolumn{2}{|c|}{\multirow{2}{*}{5 Random predefined postures}} & Decoupled & Coupled & Ours + & Ours + \\
		\multicolumn{2}{|c|}{} & Approach & Approach & Single & Multi \\
		\hline
		~ & \# of time-outs & 0 & 0 & 0 & 0 \\
		\cline{2-6}
		Problem 1 & Initial solution time & 4.79 & \multirow{2}{*}{8.88} & \multirow{2}{*}{\textbf{2.91}} & \multirow{2}{*}{4.18} \\
		Fig.~\ref{fig:scene_1} & (Base/Manipulator) & (0.54/4.25) & ~ & ~ & ~ \\
		\cline{2-6}
		~ & Final distance cost & 6.75 & 9.04 & 6.76 & \textbf{5.95} \\
		\cline{2-6}
		\hline
		\hline
		~ & \# of time-outs & 13 & 21 & 1 & 1 \\
		\cline{2-6}
		Problem 4 & Initial solution time & 41.78 & \multirow{2}{*}{55.19} & \multirow{2}{*}{17.04} & \multirow{2}{*}{\textbf{12.64}} \\
		Fig.~\ref{fig:scene_4} & (Base/Manipulator) & (1.60/40.18) & ~ & ~ & ~ \\
		\cline{2-6}
		~ & Final distance cost & 9.28 & 12.79 & 8.27 & \textbf{7.72} \\
		\cline{2-6}
		\hline
		\hline
		~ & \# of time-outs & - & 29 & 0 & 0\\
		\cline{2-6}
		Problem 6 & Initial solution time & \multirow{2}{*}{-} & \multirow{2}{*}{63.39} & \multirow{2}{*}{\textbf{42.73}} & \multirow{2}{*}{45.57} \\
		Fig.~\ref{fig:scene_6} & (Base/Manipulator) &  & ~ & ~ & ~ \\
		\cline{2-6}
		~ & Final distance cost & - & 16.13 & \textbf{14.43} & 14.65 \\
		\cline{2-6}
		\hline
	\end{tabular}
	\vspace{0.0cm}
	\label{tab:result_predefined}
\end{table}

\noindent
\textbf {Analysis with different predefined postures.}
We also test different methods with five random postures as the predefined
posture.
Table~\ref{tab:result_predefined} shows results in Problem 1, 4, and 6;
other scenes show similar tendencies.
Our methods find solutions faster than both decoupled and coupled approaches
and achieve the shortest paths in all of tested problems even with different initial postures, demonstrating the robustness of our approach. These results show that our method robustly achieves a better performance than both decoupled and coupled approaches.

\begin{figure}[t]
	\vspace{0.2cm}
	\centering 
	\includegraphics[width=2.9in]{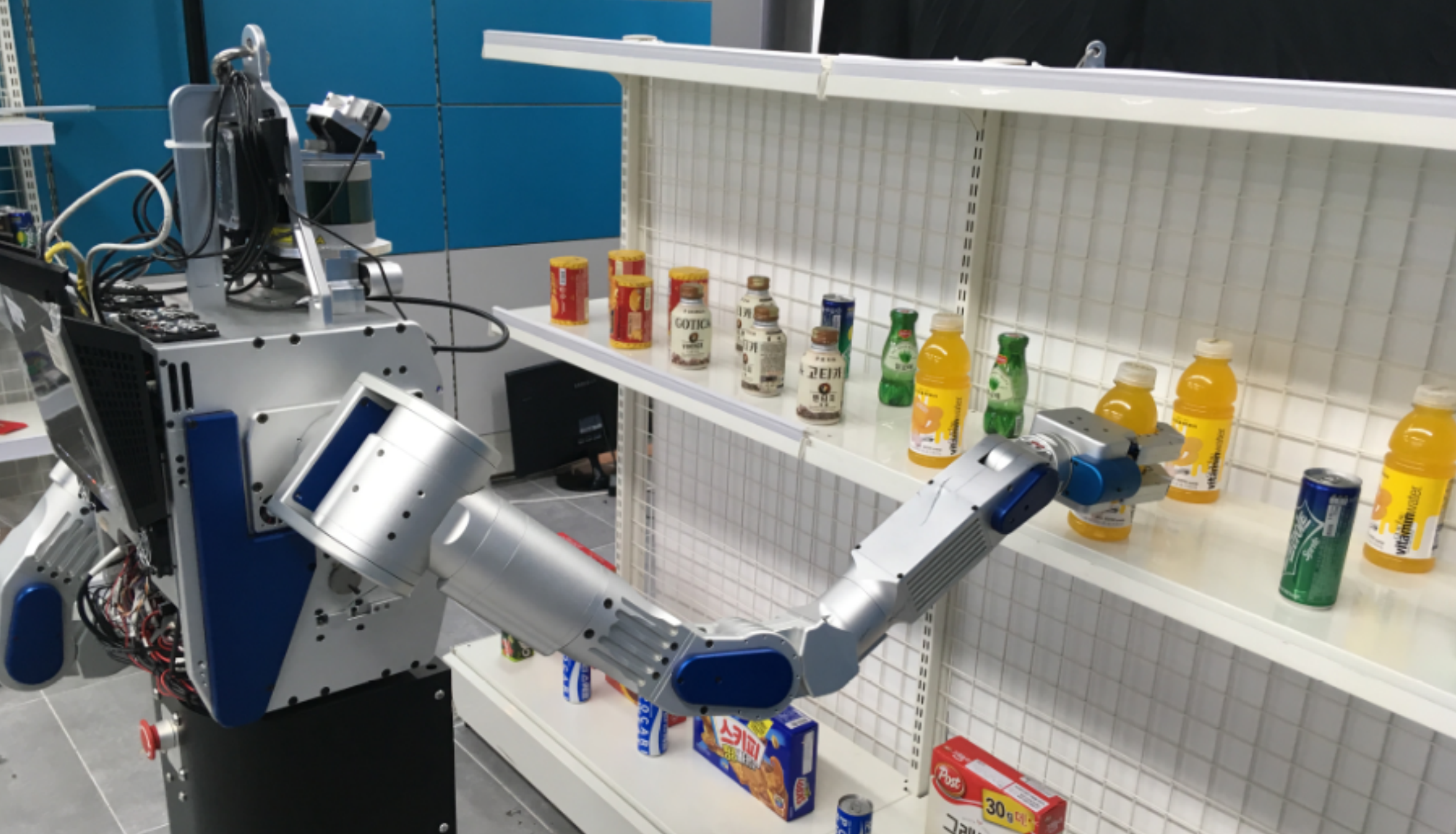}
	\caption{
		This figure shows our experimental environment  mimicking a convenient store with the mobile Hubo robot.
		There are various kinds of beverages and snacks on the shelf, 
		and our goal is to grasp the yellow beverage. 
		Our approach computes a shorter path over the decoupled
		approach in the same time budget.
	}
	\label{fig:real_experiment}
	\vspace{0.2cm}
\end{figure}

\subsection {Real Robot Test}
\label{sec:real_robot_test}

We integrated our approach with the real, mobile Hubo robot.  We tested in an
experimental environment designed for mimicking a convenience store with
various kinds of beverages and snacks on the shelf
(Fig.~\ref{fig:real_experiment}).  We defined a scenario that the robot moves to the shelf
and grasps the yellow beverage on
the shelf (Fig.~\ref{fig:main}).  In this scenario, we compared our method with the decoupled
approach.  Our method stretched out the manipulator while moving the base body.
On the other hand, the decoupled approach moves to the shelf without adjusting
the manipulator, and then adjusts the manipulator to grasp the yellow beverage,
resulting in an inferior path.
As a result, our method grasped the target object with a better solution than
the decoupled approach; i.e.,  the cost of our approach is 3.962, while the
decoupled one has 4.318.

\section{Conclusion and future work}
\label{sec:6}

In this paper, we have introduced a simple, yet effective mobile manipulation sampling method named harmonious sampling that adaptively adjusts the sampling space for the base and the manipulator. 
Our harmonious sampling works upon the low-dimensional base space to rapidly explore a wide-open space, while exploring in a high-dimensional space where it is necessary to consider full DoFs of the mobile manipulator.
Through the experiments, we have shown the improvement of performance in a
variety of indoor environments with practical tasks.  Furthermore, we have
tested our approach by integrating with the mobile Hubo in the real
environment.

In general, it is  critical to consider the uncertainty of sensing and robot
control.  Since our method
simultaneously operates for the base and manipulator, it is important to handle
the uncertainties, which are under active research in robotics.  In future
work, we would like to efficiently handle them, and support anytime
planning.

\section*{Acknowledgment}
We would like to thank anonymous reviewers for constructive comments. 
This work was supported by MI/KEIT (10070171) and NRF/MSIT (No.2019R1A2C3002833).

{
\small
\bibliographystyle{ieee/ieee}
\bibliography{./mc}
}

\end{document}